\documentclass{sig-alternate}

\usepackage{subfigure}
\usepackage{booktabs}
\usepackage{xcolor}

\newcommand\eg{\emph{e.g.}}
\newcommand\ie{\emph{i.e.}}
\newcommand\etc{\emph{etc.}}

\usepackage{algorithm}
\usepackage{algorithmic}
\usepackage{url}

\begin{document}
%

\title{Automatic Extraction of the Passing Strategies\\of Soccer Teams}

%
%
%
%
%

\numberofauthors{2} 
%
\author{
%
%
\alignauthor
Laszlo Gyarmati\\
       \affaddr{Qatar Computing Research Institute, HBKU}\\
       \email{lgyarmati@qf.org.qa}
\alignauthor
Xavier Anguera\\
       \affaddr{Telefonica Research}\\
       \email{xanguera@tid.es}
}

\maketitle
\begin{abstract}
Technology offers new ways to measure the locations of the players and of the ball in sports. This translates to the trajectories the ball takes on the field as a result of the tactics the team applies. The challenge professionals in soccer are facing is to take the reverse path: given the trajectories of the ball is it possible to infer the underlying strategy/tactic of a team? We propose a method based on Dynamic Time Warping to reveal the tactics of a team  through the analysis of repeating series of events. Based on the analysis of an entire season, we derive insights such as passing strategies for maintaining ball possession or counter attacks, and passing styles with a focus on the team or on the capabilities of the individual players. 
\end{abstract}





\section{Introduction}

Technology is reaching many sports nowadays offering new ways to improve the performance of the players to obtain competitive advantage over the opponents. In sports like soccer, American football, or basketball, the locations of the players and of the ball are measured and logs of such locations are created. This allows for analysis and visualization of the plays. For example, in soccer this translates to the trajectories the ball takes on the field, thanks to the tactics the team applies. The challenge professionals in sports are facing is to take the reverse path: given the trajectories of the ball is it possible to infer the underlying strategy/tactic of a team?  Although professionals working at soccer clubs hardly ever share insights publicly, the head of match analysis of FC Bayern Munich recently shared his view on the next steps to come in sport analytics: ``Pattern detection, repeated actions---that's what makes it a little bit more predictable and helps the team prepare''\footnote{\url{https://goo.gl/qTxEnc}}

Quantitative performance analysis in sports has become mainstream in the last decade driven by the availability of sport related datasets.
%
%
The limitation of the current systems that analyze sport-related logs is twofold. First, these systems focus on descriptive statistics that do not capture entirely the strategy of the players and teams. For example, in case of soccer, the average number of shots, goals, fouls, passes are derived both for the teams and the players~\cite{duch2010quantifying}. The systems are able to identify and evaluate the outcome of the strategies but are reluctant to extract the key components of the strategies that lead to these metered outcomes~\cite{narizuka2013statistical,lucey2013assessing,gyarmati2014}. Second, prior art does not completely address the dynamic aspect of the games, \ie, the fact that particular sequences of events might result in advantage opportunities. Moreover, during the training sessions, players are taught to apply sequences (\eg, of the passes) as part of the team's strategy. The state-of-the-art systems are not able to fully process such temporal patterns, but have shown interesting results using small databases~\cite{Gudmundsson_2012,Mutschler}. The proposed implementations are very computation intensive, making it difficult to process large trajectory logs.


Our proposal offers an advantage over these methods in terms of scalability: we are able to handle multiple leagues and seasons due to the usage of dynamic programming to identify recurring patterns (see Section~\ref{sec:algorithm} for details). We propose a method to \emph{reveal the tactics of a team }through the analysis of repeating series of events, \ie, the positions of the ball. The method is able to find fine-grained, reoccurring patterns across complete seasons, which is impossible in the traditional way. Our main contributions are:
\begin{itemize}
	\item We propose an algorithm based on Dynamic Time Warping to identify and extract the pass strategies of soccer teams automatically.
	\item We determine all the recurring pass sequences of a complete season of the Spanish first division, including patterns spanning more than 100 meters or applied as much as five times.
	\item We derive insights such as passing patterns (i) for maintaining ball possession or launching counter attacks, (ii) to enter the attacking third, and (iii) with a focus on the team's philosophy rather than the individual capabilities of the players (Section~\ref{sec:evaluation}).
\end{itemize}

Identifying recurrent pass sequences is currently done via manual analyses of the logs and of the video footage of the games by dedicated personnel of soccer teams. On the one hand, with the proposed method we can \emph{scale} such analysis by analyzing many more games that a human could manually process. On the other hand, we are able to \emph{objectively quantify} the obtained results (e.g. how often a found tactic is performed, how accurate a given sequence of movements/events is from each other, \etc).

The application of the derived insights is twofold. On the one hand, the identified passing strategies are useful in opponent scouting, \ie, in the preparation for the upcoming opponent. Our methodology reveals all the intentional (\ie, learnt at training sessions) passing strategies of the next opponent. By knowing the strategy of the opponent, a team can prepare to prevent these traits happening during the game. On the other hand, the extracted patterns provide insights on the players' involvement in these orchestrated movements, hence, it is a valuable addition to the evaluation of players (\eg, in case of player scouting). A team may prefer to sign such a player who is involved in patterns similar to those that are applied by the given team itself.

\section{Methodology}\label{sec:algorithm}
In this section, we briefly review the methodology used to extract insights by finding repeating patterns of pass events during soccer games.
The input of the system is the logs of locations on the field where passes happened. For each pass we use the information regarding the position on the field ($x$ and $y$ locations, normalized to the range $\left[0,100\right]$) both for the reception and the emission of the ball. Furthermore, we treat each team's ball possession individually, resulting in a set of two-dimensional sequences for each team for each game. An additional pre-processing step is the normalization of the passes with respect to location and time. To do so, we insert a number of virtual positions into the pass sequence such as the distance between any two positions is less than $2$ unit.





\begin{figure}[tb]
\centering
\includegraphics[width=8cm]{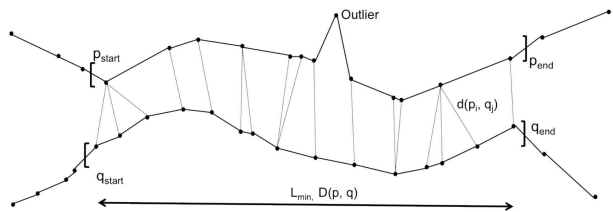}
\vspace{-3mm}
\caption{The basic idea of extracting reoccurring pass patterns using a Dynamic Time Warping (DTW) based algorithm. Two subsequences of passes match if they are close enough (both locally and globally). 
}
\label{fig:dtw_idea}
\end{figure}

We use a dynamic programming algorithm derived from Dynamic Time Warping (DTW) (like those proposed in \cite{Anguera,Minnen}) to compare sequences of passes present in the datasets. In Figure~\ref{fig:dtw_idea}, we present the idea behind the use of DTW techniques for the matching\footnote{Throughout the paper, we use the term match exclusively with respect to patterns sharing common properties, while we use game to refer to an event of soccer.} of time series, \ie, sequences of passes. We illustrate two sequences of passes that could correspond to two different sequences or to two segments in the same sequence\footnote{\ie, it may happen that a pass sequence itself has recurring parts}. Each vertex in the figure corresponds to the position of the field where a pass originates or finishes, and the straight solid line connecting them indicates the sequence order. The axes represent the $(x,y)$ coordinates of the field. We define two subsequences $p$ and $q$, with limits $(p_{start}, p_{end})$ and $(q_{start}, q_{end})$ for which we compute their optimum distance by first aligning the vertices in both subsequences and then computing their local $d(p_i, q_j)$ (between any pair of vertices) and global $D(p,q)$ distances (between the complete subsequences). We use $2$ and $10$ units as the local and global distances, respectively.

The algorithm discovers the start and end locations of the subsequences $p$ and $q$ automatically. When aligning the vertices, we allow for insertions and deletions of vertices (represented by the alignment of two or more vertices in one subsequence to the same vertex in the other subsequence). In addition, our implementation of the Dynamic Time Warping allows us to skip outliers (\ie, vertices that due to noise or other artifacts deviate from the general trend) so that these do not jeopardize the correct matching of the two subsequences. Outliers can be originated by errors in the measuring/annotation method or by small deviations from the pre-trained strategy that a player is forced to undertake due to the opponent's reaction. In addition, our method allows the application of additional constraints that ensure the parallel behavior of both trajectories as well as their minimum length ($L_{min}$). 


The output of the algorithm is the list of the most common sequences of passes, with a minimum length and similarity between them, that a particular team performs in a given game (or set of games). In our evaluations, we only consider matching sequences with more than 40 positions and at least one complete, original pass.

\begin{figure}[tb]
\centering
\includegraphics[clip=true, trim=0cm 2cm 0cm 2cm,width=4.1cm]{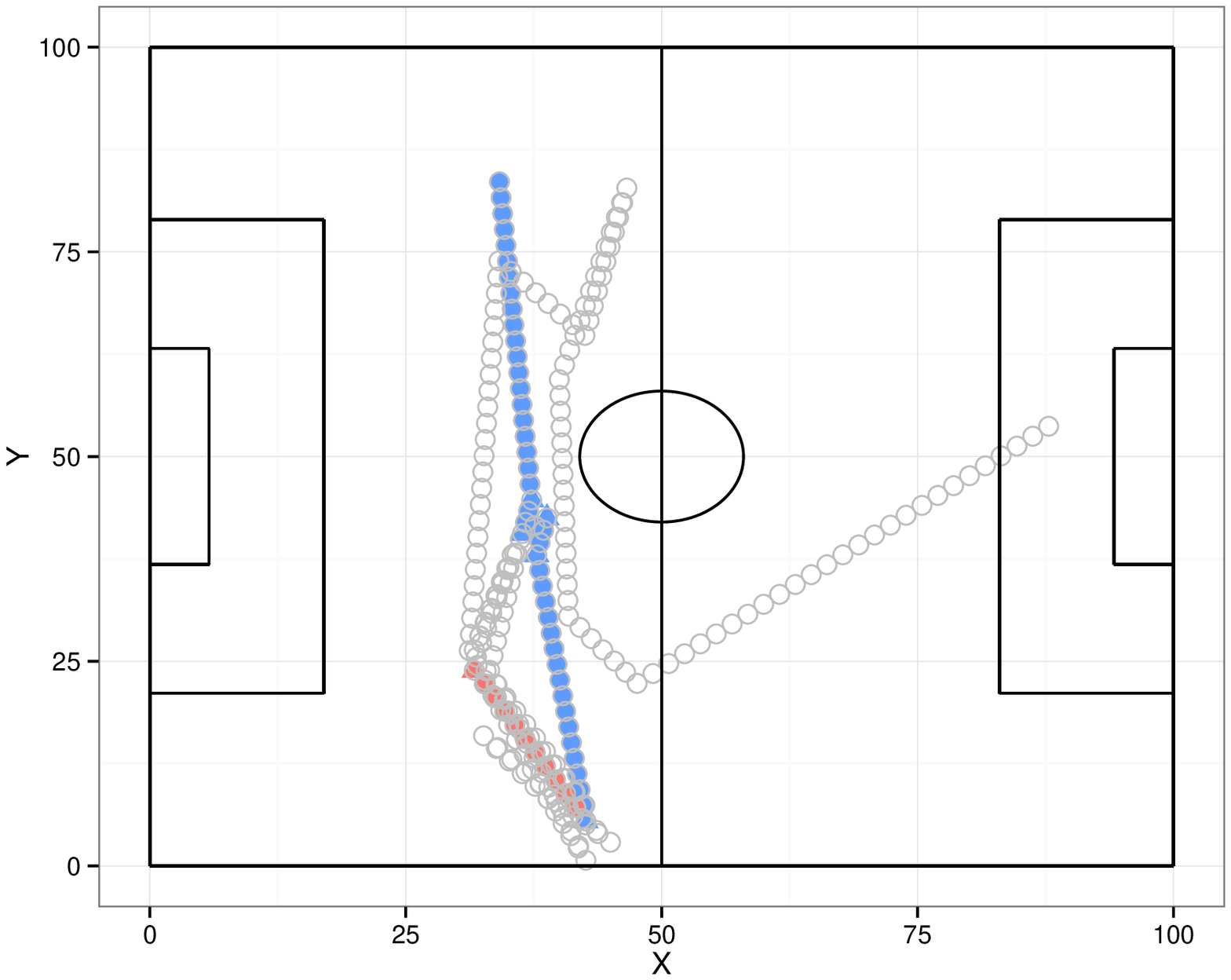}
\includegraphics[clip=true, trim=0cm 2cm 0cm 2cm,width=4.1cm]{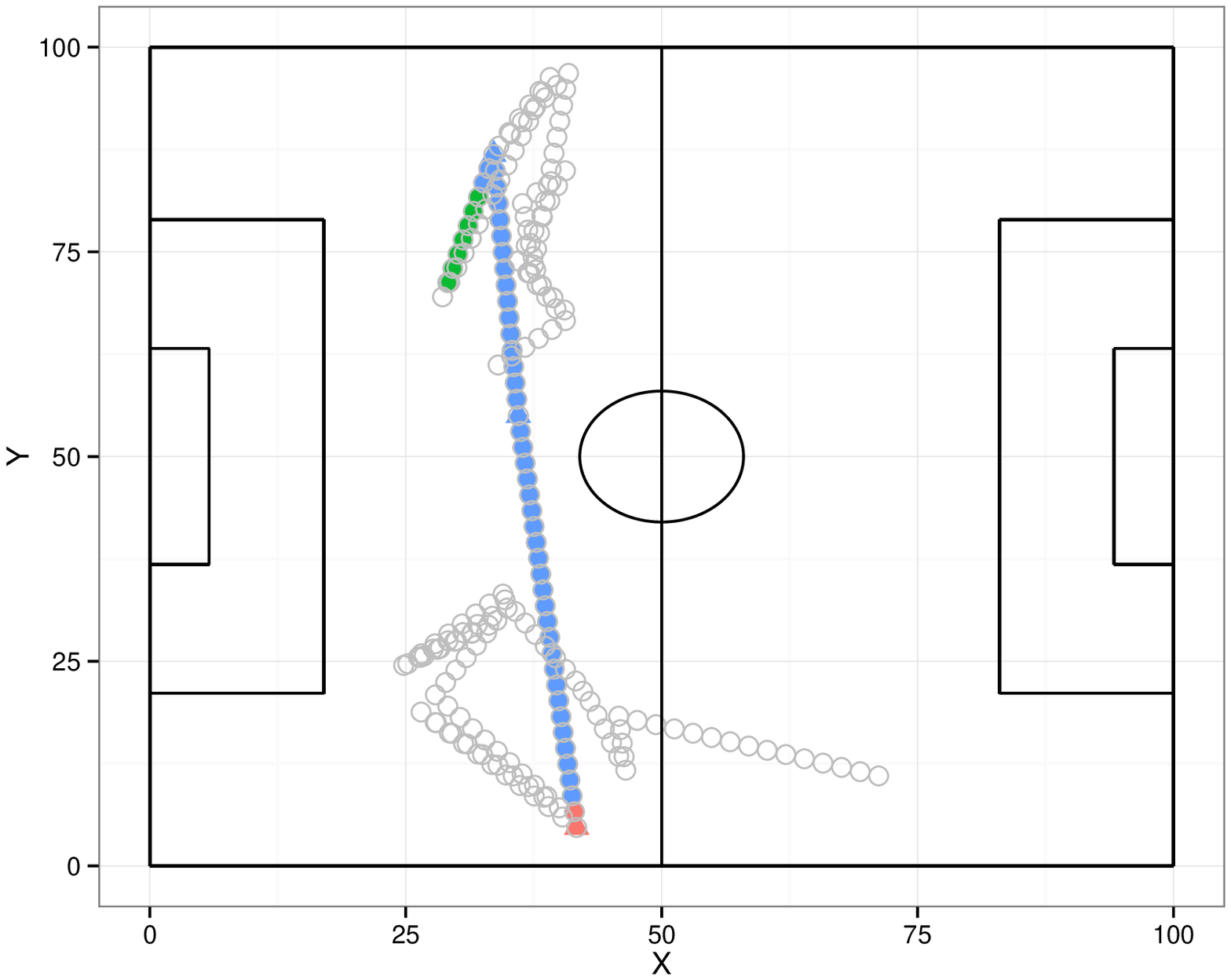}
\vspace{-4mm}
\caption{Illustration of the procedure of the proposed scheme. We extract the reoccurring pattern (blue circles) from the whole pass sequences of a team (grey circles).}
\label{fig:extraction_example}
\end{figure}

\vspace{2mm}
\noindent \textbf{Toy example.} 
We show two pass sequences of a team in Figure~\ref{fig:extraction_example}. The proposed scheme determines the parts of the sequences that share a common trajectory (denoted with blue). The pattern may not start or terminate at an endpoint of an original pass (denoted by triangles). We illustrate the pre-, and post-pattern parts of the pass sequences until reaching an endpoint of an original pass with green and red shapes, respectively. The figure also shows the resilience of our algorithm to outliers (\ie, the short back and forth passes in the middle of the reference sequence). In the remainder of the paper we denote the two instances of a pattern as reference and found sequences. This naming is just for lucidity, the algorithm itself does not apply any kind of a priori hints (\ie, references) on the patterns.

\section{Evaluation}\label{sec:evaluation}

\noindent \textbf{Dataset.}
We use the proposed method to analyze one season of the Spanish first division soccer league. Our dataset contains information on all the passes that happened in the 2013/14 season, along with the location (\ie, $(x,y)$ coordinates) of the start and end of the passes. Additionally, the dataset includes the time when the passes happened and the identity of the involved players. The dataset involves 20 teams, 380 matches, and more than 300,000 passes. We focus on the reoccurring pass patterns of individual teams, \ie, both the reference and the found pass sequences belong to the same team. This enables us to derive the strategies of the teams they follow throughout the season. 

\vspace{4mm}
\noindent \textbf{High-level statistics.}
We investigate the prevalence of designed passing sequences of individual teams, Table~\ref{tab:highlevel} presents the high-level statistics of the league. FC Barcelona and Real Madrid applied the most patterns in the season (remarkable $4.8$ and $4.0$ patterns per game), suggesting that these teams have a well-defined playing style. We stress that the difference along the joint part of the sequences are less then couple of meters. The results reveal conscious preparation for the games in case of mid-level teams such as Bilbao, Celta Vigo, and Vallecano. On the other hand, Atletico Madrid---the champion of the season---was barely using reoccurring patterns in the season.

The number of passes within the patterns shows the complexity of the passing strategy of the teams. 
Some teams, like Almeria and Espanyol, had only simple patterns: all of their extracted pass strategies had a single pass. We emphasize again that this does not mean that the matched pattern consist of only this pass, it also contains a fraction of passes and ball movements preluding and following the one complete pass. Again, FC Barcelona is the other end of the spectrum, it had the most complete passes in its patterns.

The last column of the table presents the number of final third entries enabled by the patterns, which is an important performance indicator in soccer---it shows the attacking capabilities of a team. We consider those patterns that start outside of the final third (\ie, the x coordinate of the first matching part is less then 66 units) and end within that area. Betis and Vallecano have an impressive usage rate of the patterns to enter the final third: 44.4\% and 29.5\%, respectively. FC Barcelona also applies patterns that lead into the final third frequently (25.4\%). On the contrary, only a small fraction of Real Madrid's reoccurring pass sequences are designed to enter the final third (15.1\%).

\begin{table}
	\scriptsize
	\centering
\begin{tabular}{lrrr}
\toprule
  Team &  \#PT &  \#PASS & \#FTE  \\
\midrule                                                                            
     Almeria &         27 &       1.00(0.00) &  1  \\
    Atletico Madrid &         31  &      1.10(0.30) &  10 \\
  FC Barcelona &        181 &      1.14(0.38) &  46 \\
      Betis &         27  &      1.07(0.27) &  12 \\
     Bilbao &         93  &      1.06(0.25) &  15 \\
      Celta Vigo &         99  &      1.04(0.20) &  18 \\
      Elche &         51  &      1.04(0.20) &  8 \\
   Espanyol &         17  &      1.00(0.00) &  3 \\
     Getafe &         26  &      1.12(0.33) &  9 \\
    Granada &         79  &      1.06(0.25) &  10 \\
    Levante &         19  &      1.11(0.33) &  2 \\
      Malaga &         44 &      1.16(0.37) &  10 \\
    Osasuna &         53 &      1.06(0.23) &  10 \\
   Real Madrid &        152 &      1.08(0.29) &  23 \\
    Sevilla &         51  &      1.08(0.27) &  5 \\
   Sociedad &         89  &      1.06(0.28) &  17 \\
   Valencia &         63  &      1.06(0.25) &  11 \\
 Valladolid &         52  &      1.12(0.32) &  11 \\
  Vallecano &        105  &      1.08(0.27) &  31 \\
 Villarreal &         39  &      1.10(0.31) &  0 \\
\bottomrule
\end{tabular}
\vspace{-1mm}
\caption{High-level statistics of the matching patterns: number of patterns (PT), the average number of passes within (and their std.dev.), and the number of final third entries using the patterns (FTE).}
\label{tab:highlevel}
\end{table}

Analyzing the time duration of the patterns reveals that 42\% of the patterns are 3--5 seconds long. However, in some extreme cases, a pattern may take as much as 25 seconds. 
In terms of the length, 68 percent of the patterns are less than 60 meters long, however, we managed to extract 12 patterns with length as long as 100 meters. There is a strong connection between the number of passes and reoccurring patters a team has during a season ($R^2=0.83$), \ie, more passes lead to more matched patterns. The two teams most apart from this trend are from Madrid: Real applies the patterns more frequently than expected, while Atletico has way fewer patterns than it supposed to have.

\begin{figure*}[tb]
\centering
\includegraphics[width=14cm]{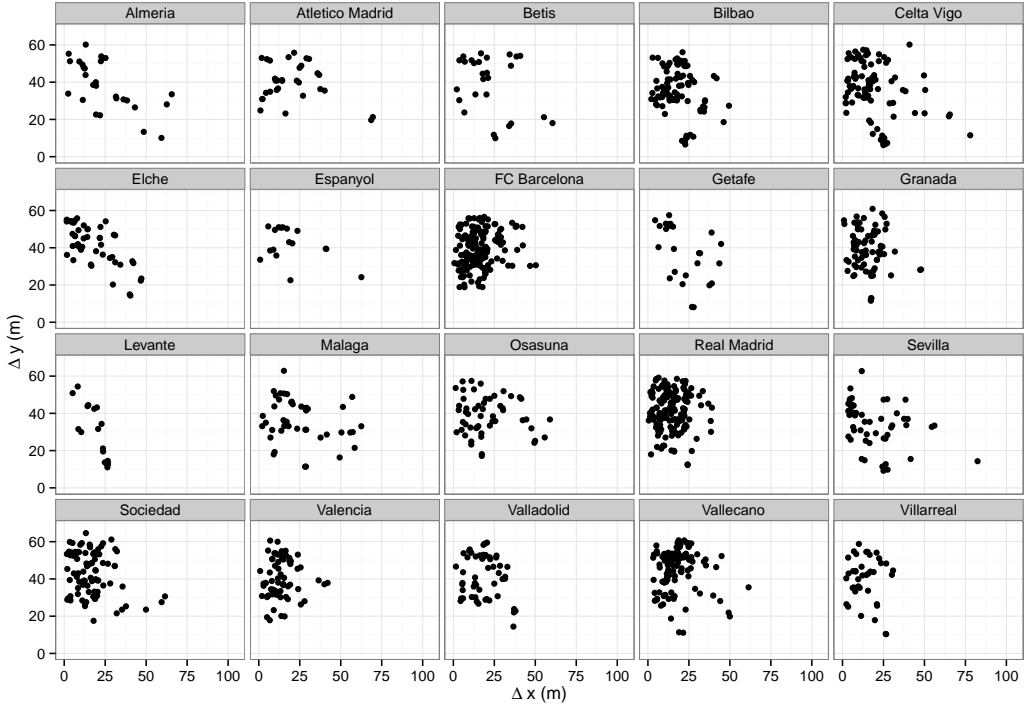}
\vspace{-6mm}
\caption{Movements during the patterns. The analysis of the recurrent patterns reveal the underlying strategies of the teams: counter attacks with (\eg, Malaga) and without (\eg, Atletico Madrid) side changes, balanced ball possession (\eg, Real Madrid and FC Barcelona), and crosses (\eg, Vallecano).}
\label{fig:movement_delta}
\end{figure*}

\vspace{3mm}
\noindent \textbf{Team strategies.}
We next give more attention to the individual strategies of the teams. 
The spatial spread of the patterns contains insight not only on the length of the patters, but also reveals the aim of the teams. A team with patterns of high vertical span focuses on retaining the ball, while large horizontal pattern are useful for attacking purposes. We define the spatial spread of the patterns as the difference of the $x$ and $y$ coordinates of the first and last part of the matched pattern. We illustrate the spatial variation of the teams' patterns in Figure~\ref{fig:movement_delta}. The results show several distinct strategies. First, there are teams that have patterns for direct counter attacks. Teams like Atletico Madrid, Sevilla, Celta Vigo, \etc, have patterns that move the ball 60+ meters closer to the opponent's goal while keeping the ball at the same side of the pitch. Second, Malaga has also counter attack focused patterns: large (40-60 meters) differences in the $x$ coordinate, however, it also changes the side of the field (the vertical change is larger than half of the field). A third group of teams is more focused on having balanced reoccurring patterns, the horizontal span of 20-30 meters are combined with large variety of vertical dilatations. This group consists of FC Barcelona, Real Madrid, and Valencia. The last pass strategy involves usage of crosses with modest change in the $x$ coordinate (\eg, Vallecano). 

\begin{figure}[tb]
\centering
\includegraphics[width=6cm]{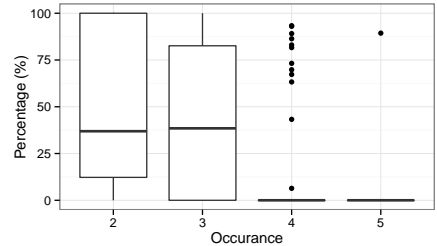}
\vspace{-5mm}
\caption{Some patterns were applied more than two times during the season. The percentage of the overlapping parts reveals the similarity of the patterns with multiple occurrences. 
}
\label{fig:multiple_matching}
\end{figure}

We next address the following question: do such patterns exist that occur multiple times? 
The greedy nature of the method maximizes the length of the matched sequences under the defined similarity constraints, thus, exact matching of multiple sequences are highly unlikely. However, the patterns may have overlaps. To quantify this, we compute for all the parts of the patterns how many times they are members of the overlapping patterns. Hence, we have a percentage of overlap for the given number of occurrences. Figure~\ref{fig:multiple_matching} reveals these overlapping percentages and the number of the pattern occurrences. There is only a single pattern that happened 5-times during the season (the occurrences have a 90\% overlap ratio). This pattern, which is basically a cross from the left side to the right side, was used by Vallecano. 
Multiple teams have patterns with four occurrences including FC Barcelona, Bilbao, Levante, Real Madrid, and Sevilla.

Are patterns a result of a team-focused strategy of the team (in case the players are different in the matched sequences) or some players are crucial for the strategy (in case the players are identical or there is a significant overlap between the set of involved players). To answer this, we focus on the players' involvement in the patterns. Specifically, we determine the set of players who are involved in the reference and the found sequences, and then we compute the ratio of the overlapping players. Table~\ref{tab:player_overlap} presents the number of patterns with the given number of involved and overlapped players. In aggregate, there are 83 patterns in which the involved players are identical. These patterns include one, two, or three players. 30 patterns incorporate 4 different players, however, none of these patterns have complete overlap of the players. If we focus on the distributions of the overlapping ratios, three types of teams arise (figure not shown).
The first group of teams use mostly team strategies, \ie, the overlap of the patterns' players are marginal. Teams like Almeria and Sevilla are belonging to this group. The second group---with team such as FC Barcelona, Real Madrid, or Betis---uses more player specific strategies. A significant portion of the patterns have 50 percent overlap in terms of the involved players. The third class of the teams uses a strategy that is the mixture of the above two: both player and team strategies are prevalent. Elche and Atletico Madrid are examples of this philosophy.

\begin{table}
	\scriptsize
	\centering
\begin{tabular}{l|rrrrr}
& \multicolumn{5}{ c }{\#overlaps} \\
\# players &    0 &    1 &   2 &   3 & 4 \\
\midrule
1                        &   48 &    9 & - & - & -\\
2                        &  449 &  297 &  62 & - & -\\
3                        &  143 &  172 &  70 &  12 & - \\
4                        &   12 &   11 &   5 &   2 & 0 \\
\end{tabular}
\vspace{-3mm}
\caption{The number of matching patterns with the given number of involved and overlapping players. Patterns using exactly the same set of players are good indication of a player-focused passing strategy.}
\label{tab:player_overlap}
\end{table}

\vspace{4mm}
\section{Conclusions} \label{sec:conclusions}

Building on a DTW based algorithm, we are able to extract all the recurring pass sequences of whole soccer seasons automatically, which is, based on our numerous discussions with domain experts, unprecedented and innovative in the soccer industry. The identified patterns reveal novel insights on the playing style and strategy of the teams that can be used as a competitive advantage.
As future work, we plan to analyze multiple, consecutive seasons to study the stability of the teams' applied patterns. 

\bibliographystyle{abbrv}
\small{
\vspace{1mm}
\bibliography{sigproc}  
}
%
%
\end{document}